\documentclass{ieeeaccess}
\usepackage{cite}
\usepackage{amsmath,amssymb,amsfonts}
\usepackage{algorithmic}
\usepackage{graphicx}
\usepackage{textcomp}
\usepackage{arydshln}
\usepackage{color}
\usepackage{caption}

\newcommand{\highlight}[1]{{#1}}

\DeclareCaptionFont{ieeeblue}{\color{accessblue}}
\DeclareCaptionLabelFormat{myformat}{\figcapfont{\textbf{#1}\textbf{#2}}}
\def\BibTeX{{\rm B\kern-.05em{\sc i\kern-.025em b}\kern-.08em
    T\kern-.1667em\lower.7ex\hbox{E}\kern-.125emX}}
\begin{document}
\history{Date of publication xxxx 00, 0000, date of current version xxxx 00, 0000.}
\doi{10.1109/ACCESS.2019.2947090}

\title{Sequential Image-based Attention Network for Inferring Force Estimation without Haptic Sensor}
\author{\uppercase{Hochul Shin\authorrefmark{1}, Hyeon Cho\authorrefmark{1}, Dongyi Kim\authorrefmark{1}},
\uppercase{Dae-kwan Ko\authorrefmark{2}, Soo-Chul Lim\authorrefmark{2}}, \IEEEmembership{Member, IEEE},
\uppercase{and Wonjun Hwang}.\authorrefmark{1},\IEEEmembership{Member, IEEE}}
\address[1]{Dept. of Software and Computer Engineering, Ajou University, San 5-1, Woncheon-dong, Yeongtong-gu, Suwon-si, Gyeonggi-do, 16499, Republic of Korea (e-mail: hochul.shin@nhn.com, ch0104@ajou.ac.kr, waps12b@ajou.ac.kr, wjhwang@ajou.ac.kr)}
\address[2]{Dongguk University, Republic of Korea (e-mail: kodaekwan@dongguk.edu, limsc@dongguk.edu)}
\tfootnote{This work was supported by the Samsung Research Funding Center of Samsung Electronics under Project SRFC-TB1703-02.}

\markboth
{H. Shin \headeretal: Sequential Image-based Attention Network for Inferring Force Estimation without Haptic Sensor}
{H. Shin \headeretal: Sequential Image-based Attention Network for Inferring Force Estimation without Haptic Sensor}

\corresp{Corresponding author: Wonjun Hwang and Soo-Chul Lim (e-mail: wjhwang@ajou.ac.kr and limsc@dongguk.edu).}

\begin{abstract}
Humans can approximately infer the force of interaction between objects using only visual information because we have learned it through experiences. Based on this idea, in this paper, we propose a method based on a recurrent convolutional neural network that uses sequential images to infer the interaction force without using a haptic sensor. To train and validate deep learning methods, we collected a large number of images and corresponding data concerning the interaction forces between objects shown therein through an electronic motor-based device. To focus on the changing appearances of a target object owing to external force in the images, we develop a sequential image-based attention module that learns a salient model from temporal dynamics for predicting unknown interaction forces. We propose a sequential image-based spatial attention module and a sequential image-based channel attention module, which are extended to exploit multiple images based on corresponding weighted average pooling layers. Extensive experimental results verified that the proposed method can successfully infer interaction forces in various conditions featuring different target materials, changes in illumination, and directions of external forces.
\end{abstract}


\titlepgskip=-15pt

\maketitle

\section{Introduction}
\PARstart{O}{f} the five basic human senses, touch is an important perceptional modality for understanding the relationship between our surroundings and us. It offers complementary information that helps comprehend the environment. From this perspective, touch or tactile sensing has been an attractive subject of research in robotics and haptics for many years \cite{h1}\cite{h2}\cite{h3}\cite{haptic2016}\cite{haptic2017}. The main physical property needed for grasping and interacting with objects is the force of interaction. When a robotic hand attempts to grasp an object, a contact-type haptic sensor is used to measure the force of interaction between the device and the object. This improves the success rate of the gripping and enables precise hand manipulations~\cite{Lim2015}. In the case of humans, the visual information sensed through the eyes is used in addition to tactile sensations when grasping an object. Through visual information, we perceive the shape, appearance, and texture of an object, and infer the tactile memory learned through the past experience. From the perspectives of neuroscience and psychophysics, Ernst and Banks~\cite{ref1} investigated the mechanism of information sharing between the senses of vision and touch. Newell et al.~\cite{ref3} showed that the human brain employs shared models of objects across multiple sensory modalities, e.g., vision and tactile sensing, so that knowledge can be transferred from one to another.

\begin{figure}
    \begin{center}
    \includegraphics[width=8.0cm]{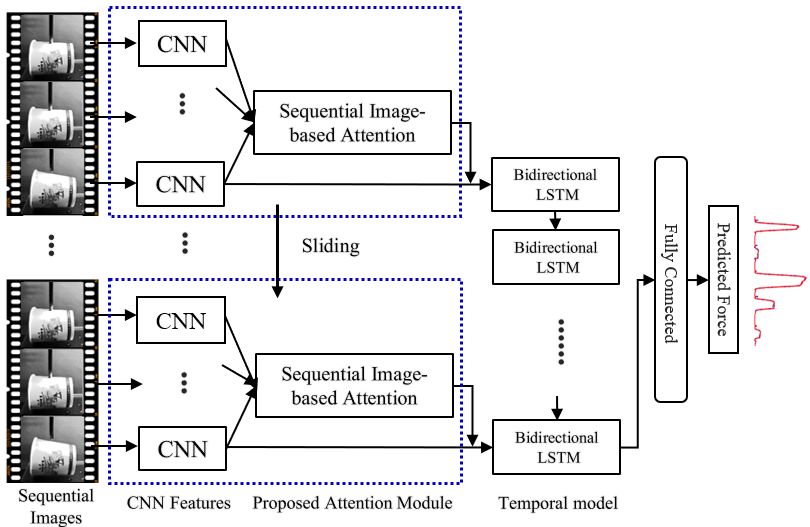}
    \end{center}
    \caption{Predicting haptic force based on the proposed sequential image-based attention module using only images.}
    \label{fig:1}
\end{figure}

Inspired by the knowledge transfer from vision to touch~\cite{Tiest2014}, we propose a vision sensor-based method that simulates tactile sensing, which has a different modality, using only visual sensing information. When humans try to touch an object, they can recall how it feels before touching it by summoning past experiences. Specifically, if we know what an object is, and we can observe how its appearance changes by a finger pressing, we can predict the interaction force between the object and the finger from past experiences. Another focus of the proposed method is that compared with contact-type haptic sensors, a non-contact-type sensing method can help constantly measure the haptic force because the camera sensor is not worn out even when it is thus used for a long time. Moreover, as an additional touch sensor does not need to be attached to the instrument, the mechanism of the instrument can be miniaturized.
In this paper, our computational approach is based on learning haptic information from past human experiences. The following are pivotal rules: to predict the interaction forces exerted by the target object using only sequential images, and for simplicity, we assume that the target objects have been touched in advance like past experiences.
%(1) to recognize what an object using only images, and (2) to predict the kind of interaction force exerted by the object using sequential images.
For this purpose, we collected more than 300,000 images of different objects under a variety of conditions and used the corresponding databases to train and validate the proposed method.

From the viewpoint of predicting haptic information from images, the basic deep learning architecture is developed using a convolutional neural network (CNN)-based recurrent neural network (RNN)~\cite{PAMI2017}. As in human perception processes, we used the CNN to analyze types of target objects and changes in their appearances using the images, analyzed the images over time, and used temporal changes in them as inputs to the RNN to eventually estimate the force of interaction.
To construct the network, we believe that the attention mechanism~\cite{ReA}\cite{CBA}, which focuses only on regions of importance in images for visual question answering (VQA)~\cite{VQA}, helps improve the accuracy of prediction of the force. However, the main difference between the proposed method and previously developed attention networks~\cite{ReA}\cite{CBA}, which commonly have been designed for a single image-based attention mechanism, is that we use a temporal dynamics-based attention method using sequential images to predict the interaction forces. Because the appearance changes of the target object between the sequential images play a pivotal role in inferring haptic information.
% 영상이 늘어남에 따라 생기는 정보량을 효율적으로 처리하기 위함
As the number of CNN features increase due to the sequential images, there is an increasing need for a method to efficiently process a large amount of information generated.
For this purpose, we propose a sequential image-based attention method consisting of a sequential spatial attention module (SSAM) and a sequential image-based channel attention module (SCAM) to attain higher accuracy for predicting haptic information. By developing the attention module based on sequential images independently of the RNN, as shown in Fig.~\ref{fig:1}, the concentrated information can be inferred clearly to predict the haptic force based on changes in the appearance of the target object. Moreover, we use spatial and channel attention modules, respectively, and each attention module is based on a proposed weighted average pooling (WAP) method for handling successfully a large amount of information generated by the sequential images.
%Unlike in~\cite{CBA}, we train the SSAM and SCAM independently and then merge them. Spatial and channel-related data are mutually exclusive, and are thus challenging to train under a unified framework for predicting haptic forces using sequential images.

The main contributions of this paper are as follows: (1) We propose a computational method for predicting the haptic interaction force only from visual information without a haptic sensor. (2) The sequential image-based attention modules are proposed for efficiently processing the increased convolutional features due to the sequential images and for obtaining more accurate haptic information at the same time. (3) We collected a large number of sequential images of objects and the corresponding information concerning the forces of interaction on the objects by using an automatic mechanism.

%-------------------------------------------------------------------------
\section{Related Work}
Studies have been conducted to measure interaction forces without force sensors. In~\cite{Ref22}, a stereo camera was used to reconstruct a 3D artificial heart surface and a supervised learning method was applied to predict the applied force. In \cite{Ref24}, a video-based method to estimate the interaction force between a human body and an object was proposed using 3D modeling information. In~\cite{Ref25}, a single RGB-D camera-based method was used to estimate the contact forces between a human hand and an object. %The method makes use of only visual information, given the geometrical and physical properties of the object.
In~\cite{Maryland}, a deep learning-based hand action prediction method was proposed using only visual information. It can predict the force exerted by the fingertips using the proposed networks. In~\cite{Sensor}, the authors focused on predicting the interaction force using visual changes to the target objects by using a simple RNN-based method. Their work is the first to focus on predicting the interaction force using only images without additional sensors. However, the proposed RNN-based method does not have deeper layers to effectively train all variations in visual changes, such as illumination and pose changes, at the same time. To solve this problem, we employ the basic framework of the CNN-based RNN method in which the CNN first analyzes variations in salient visual features using the proposed sequential image-based attention module, and the RNN works on the serialized features to predict the final interaction force.
%Our proposed method can thus attain more robust accuracy with respect to variations in conventional images, such as different objects, various illumination conditions, and camera pose variations.
In this respect, compared with the previous work~\cite{Sensor}, our proposed method can thus train deep learning network successfully using the various images and show better accuracy in predicting the interaction force from the sequential images.

%%
%% Attention Related Works
%%
From the viewpoint of the attention-based networks, CNN-based attention mechanism has been widely studied such as image caption generation~\cite{Bengio2015}, image classification~\cite{Zheng2017}\cite{CBA}, and visual sentiment analysis~\cite{You2017}.
\highlight{Self-attention is a novel attention mechanism relating different positions of a single sequence in order to compute a representation of the sequence. Self-attention has proved its effectiveness in a variety of fields~\cite{Self1}\cite{Self2}. In the field of classification and detection tasks, self-attention also contributed to outperforms the baseline of the convolutional neural networks~\cite{Self3}. In this paper, we have further interests to how the attention method could be applied to the image-based haptic interaction estimation. }
In case of action recognition, in~\cite{Liu2017}, LSTM-based attention was proposed to learn the attention weight between the LSTM, and the global long sequence attentive network~\cite{Yang2018} was proposed, designed on the spatial attention based on sub-sequence attention network using the sub-skeleton images.
In~\cite{Xu2017}, the attentive spatial-temporal pooling was proposed for video-based person re-identification, where they use the similarity scores of two videos to compute attention vectors and the attention vectors were used to perform pooling after RNN outputs.
% 대부분의 경우는 LSTM에서 처리한다.
% temporal attention이면서도 CNN에서 처리한다가 다르다.
Most of the temporal attentions for a video-based recognition have been developed for improving the baseline of LSTM, not CNN by itself, but in this paper, we propose the sequential image-based attention method for enhancing the performance of the convolutional features.

\section{Proposed Method}
\subsection{Baseline Method}
\label{baseline}
%\section{Baseline method: CNN-based LSTM method for sequential image description~\cite{PAMI2017}}
The baseline algorithm~\cite{PAMI2017} consists of CNN (visual feature extraction) and RNN (temporal dynamics modeling) and is described as follows.

\textbf{Visual Feature Extraction}
The CNN is recently a well-known method for a representation of images.
In case of sequential data, each frame could be represented by its corresponding feature of the CNN.
The $t$th image frame passes through visual extractor $V$ as an input $I_t$, and the CNN generates the fixed-length visual feature vector representation: $X_t=V(I_t)$. To confirm the feasibility of our model, we use a variant of the VGG model \cite{VGG} as an encoder%, and then prove generality by using the ResNet model \cite{ResNet}
, which is a common deep CNN architecture. We extract feature maps from the last pooling layer. The convolutional features of each frame are considered one chunk for an input step of the RNN. The resulting frame-level vector is fed into our long short-term memory (LSTM) architecture.

\textbf{Sequential LSTM Model}
Given a frame-level feature vector $X_t$ in sequential frames, we use $X_t$ as input to the LSTM, which is known to perform well on many sequential problems~\cite{PAMI2017}\cite{ICME2016}. To extract sequential features, we apply an LSTM comprising self-recurrent units and a memory cell. It can store information concerning several dozen time steps. We use the bidirectional LSTM (BLSTM)~\cite{ICASSP2013} derived from the LSTM. It considers all available information concerning the past and the future. As the BLSTM uses inputs in two ways, i.e., from past to future, and from future to past, there are two hidden-state outputs. We combine them in the last time step and send them to a fully connected layer.

\subsection{Sequential Image-based Attention Module for inferring haptic information}
\begin{figure*}
    \begin{center}
    \includegraphics[width=13cm]{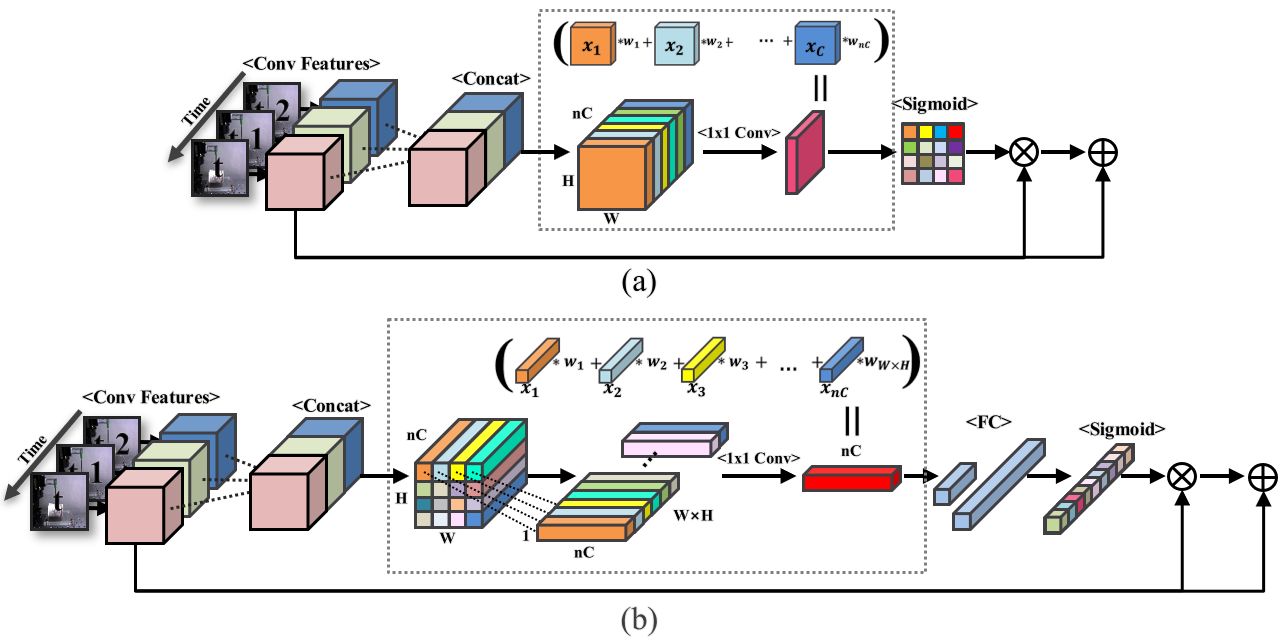}
    \end{center}
    \caption{Illustration of the proposed attention network architecture for (a) the sequential image-based spatial attention module, and (b) the sequential image-based channel attention module. Dotted boxes are the proposed WAP for the spatial and channel information, respectively. }
    \label{fig:3}
\end{figure*}
In this section, we describe the dynamic attention module designed to model the interaction between the objects by using the sequential images. As described in Section~\ref{baseline}, we used the CNN-based RNN module as the baseline for analyzing sequential images. The CNN first extracts the visual features of each frame that are passed to the RNN to predict the interaction forces based on complex temporal dynamics. The sequential attention module focuses on salient regions and considers temporal dynamic information simultaneously, as illustrated in Fig.~\ref{fig:1}.

\textbf{Sequential Image-based Spatial Attention Module (SSAM)}
In general, an interaction between objects occurs in the region that is touched; therefore, the application of a global image feature may lead to a sub-optimal result owing to its consideration of irrelevant regions. To solve this problem, a spatial attention mechanism has been proposed in many previous works~\cite{VQA}\cite{ReA}\cite{CBA}. Such a mechanism focuses on the key regions of information in an image by excluding less important ones, and has yielded improvements in performance. 
%However, most previous works~\cite{VQA}\cite{Zheng2017}\cite{You2017} have assumed that only a single frame is used for the convolutional features or the temporal attention methods~\cite{Liu2017}\cite{Xu2017} have been proposed for the RNN. 
From a single image, the convolutional attention map is inferred for the specific purpose~\cite{VQA}\cite{Zheng2017}\cite{You2017}. On the other hand, RNN-based attention~\cite{Liu2017}\cite{Xu2017} is designed to avoid the vanishment of the temporal information. 
As the purpose of this work is to predict the interaction force between objects in the sequential images, the consideration of the dynamic information of each convolutional feature is also important. Therefore, instead of extracting only an attention map from an individual frame, our attention module exploits the multiple adjacent frames to generate an accurate attention map by considering dynamic information for the convolutional features. The overall procedure is illustrated in Fig.~\ref{fig:3} (a).

We basically represent the convolutional feature of the $t$th frame as $X_t\in \Re^{H\times W\times C}$. Unlike the existing methods~\cite{VQA}\cite{ReA}\cite{CBA}, we make use of the previous frames for extracting more salient attention information. For predicting the interaction forces using the camera, it is important how the appearance of an object changes in the sequential images rather than using only one image. In this respect, the sequential convolutional feature of the $n$ sequential frames are concatenated as $X_{tn}=[X_{(t-n+1)},...,X_t ]$ in $X_{tn}\in \Re^{H\times W\times nC}$.
The SSAM process can be summarized as follows:
\begin{equation}
	X_t^{'} = M_s (X_{tn})\otimes X_t + X_t,
	\label{eq:1}
\end{equation}
where $\otimes$ denotes the element-wise multiplication operation, $M_s\in \Re^{H\times W\times 1}$ represents the sequential image-based spatial attention map, and $X_t^{'} \in \Re^{H\times W\times C}$ is the spatial-wise excited feature map.
\begin{equation}
	M_s (X_{tn}) = \sigma(\omega_s*X_{tn}),
	\label{eq:2}
\end{equation}
where $*$ denotes the convolution operation and $X_{tn}\in \Re^{H\times W\times nC}$ represents concatenated convolutional features from the $(t-n+1)$th to the $t$th image.
To squeeze the concatenated feature map $X_{tn}$ by using the proposed WAP for spatial information,
we use a 1$\times$1 convolution kernel $\omega_s\in \Re^{1\times 1\times nC}$ to generate projection tensor $Y_s\in \Re^{H\times W}$. Each $y_{i,j}$ of $Y$ represents a linear combination of all $C$ channels at spatial location $(i,j)$. To generate an attention map, the projected map $Y$ passes the convolution layer and the sigmoid function is applied as follows:
\begin{equation}
	M_s=\sigma(Y_s+b),
	\label{eq:3}
\end{equation}
where $\sigma$ is the sigmoid function, and $b$ is the bias parameter.

\textbf{Sequential Image-based Channel Attention Module (SCAM)}
Similar to the SSAM, the proposed SCAM generates salient features by exploiting the channel information of frames adjacent to the given one. As the amount of channel information increases because of multiple images, so does redundant channel information. In this case, as noted in \cite{PGA}, non-salient channel information causes the problem of distraction. To solve this issue, we use the self-gating attention module based on channel dependence \cite{SE} in the proposed channel-wised WAP method. Fig.~\ref{fig:3} (b) describes the overall block architecture of SCAM.

The set of visual features of sequential frames $X_{tn}=[X_{(t-n+1)},...,X_t ]$ is given as input as follows:
\begin{equation}
	X_t^{'}  = M_c (X_{tn} )\otimes X_t + X_t,
	\label{eq:4}
\end{equation}
where %$\otimes$ denotes the element-wise multiplication operation,
$M_c\in \Re^{1\times 1\times nC}$ represents the sequential channel attention map, and $X_t^{'} \in \Re^{H\times W\times C}$ is the final refined feature map,
\begin{equation}
	Y_c= \omega_c*reshape(X_{tn}).
	\label{eq:5}
\end{equation}

To squeeze the concatenated feature map $X_{tn}$ into the channel axis, we use 1$\times$1 convolution kernel $\omega_c\in \Re^{1\times 1\times (H\cdot W)}$ after reshaping $X_{tn}$ to obtain squeezed vector $Y_c\in \Re^{1\times nC \times 1}$. Each $y_k$ of $Y_c$ represents the linear combination of all spatial positions in channel $k$. The output then passes through two MLP layers to provide non-linear dependencies, and the sigmoid function is then applied as follows:
\begin{equation}
	M_c=\sigma (F_1 (F_0 (Y_c))),
	\label{eq:6}
\end{equation}
where %$\sigma$ is the sigmoid function,
$F_0\in \Re^{(nC/r)\times nC}$ and $F_1\in \Re^{nC\times (nC/r)}$ are the parameter weights of the multilayer perceptron, and $r$ is the reduction ratio.

%-------------------------------------------------------------------------
\textbf{Weighted Average Pooling (WAP)}
%\begin{figure}
%    \begin{center}
%    \includegraphics[width=8.0cm]{./pic/fig2.jpg}
%    \end{center}
%    \caption{Weighted average pooling (WAP) of (a) spatial information, where channel information is averaged using a 1$\times$1 convolutional layer; (b) WAP for channel information, where spatial information is averaged after reshaping tensors and applying a 1$\times$1 convolutional layer. }
%    \label{fig:2}
%\end{figure}
Recent works~\cite{ResNet}\cite{SE} have used the Global Average Pooling (GAP) to calculate the spatial average of the convolutional feature map and this type of pooling helps to achieve better accuracy in visual recognition. GAP can be used to efficiently encode several convolutional feature maps into a vector with a limited size. Therefore, many attention methods~\cite{SE}\cite{CBA} have widely used GAP due to its simplicity and efficiency.
% to extract the salient feature vector to predict regions of interest. However, this method uses equal weighted average pooling to reduce dimensionality because due to its simplicity and efficiency.
However, we argue for this simple assumption and propose the WAP method which could encode the convolutional features with consideration of their importance for spatial and channel attentions, respectively. In this paper, the proposed attention method for CNN features makes use of the increased number of feature information due to the sequential images and it is not an easy task to calculate the feature information with the same weights as GAP. For example, if the channel attention is obtained using two frames, the channel information extracted from the current frame should be calculated with higher weight than the information extracted from the previous frame.
%Moreover, we use multiple images for developing a temporal dynamics-based attention mechanism for CNN feature extraction. Compared with the single image-based method, the size of the convolutional feature map of a method that uses multiple images generally increases owing to redundant information.
Therefore, the proposed WAP encourages the network to emphasize more discriminative features when the feature information is increased using the sequential images.

As shown in Fig.~\ref{fig:3} (a), to average the channel information by using different weights, the convolutional feature $X\in \Re^{H\times W\times nC}$ is split into $\{x_1,x_2,...,x_{nC}\}$ ($x\in \Re^{H\times W}$). We calculate the weighted average by multiplying each element of weight vector $w\in \Re^{nC}$ to the corresponding spatial map. In this respect, we simply implement it by applying a 1$\times$1 convolution operation. A similar approach can be used to average spatial information using different weights as shown in Fig.~\ref{fig:3} (b). In this case, we flatten the tensors of the convolutional feature maps, e.g., $X\in \Re^{H\times W\times nC}\rightarrow \Re^{1\times nC\times (H\cdot W)}$, and apply a 1$\times$1 convolution to obtain different weights for the spatial regions.

\subsection{Ensemble Module}
The ensemble network has shown better accuracy in many applications~\cite{DeepFace}\cite{ImageNet}. To combine the attention networks, Woo et al.~\cite{CBA} designed serialized spatial and channel-wise attention modules under a single network. However, in this study, we trained the SSAM and SCAM independently and calculated the average of their results based on the late fusion rule. A major reason for this merging using late fusion is that the two proposed attention mechanisms play different roles and focus on different characteristics to infer the forces. SSAM focuses on the spatial regions in images, whereas SCAM is responsible for evaluating which channels of the convolution layer are important. Learning the two attention methods, SSAM and SCAM, the characteristics of which are different under a single network, is challenging. Moreover, we used multiple images to learn more temporal dynamics for better performance. The amount of information to be assessed by the proposed method increased compared with that in the single image-based attention method, and separately training the SSAM and the SCAM is a better choice in terms of efficiency.

\section{Dataset and Implementation}

\subsection{Experimental Setup and Database}
\begin{figure*}
    \begin{center}
    \includegraphics[width=15.0cm]{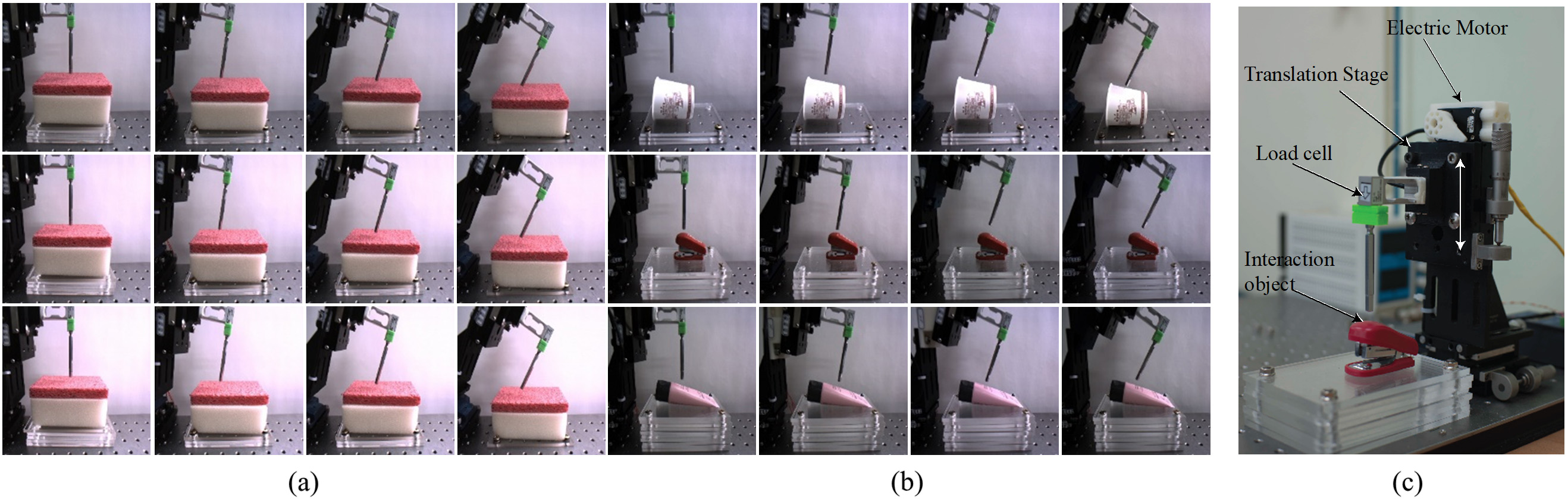}
    \end{center}
    \caption{Dataset collected to estimate interaction forces. It consisted of four objects: a sponge, a paper cup, a tube, and a stapler. An external force was applied at four pressing angles and three illumination changes occurred. (a) Images of sample images of the sponge according to each condition and variation in pressing angle, and (b) examples of images of the paper cup, the tube, and the stapler. All images and their corresponding forces were collected using (c) the data collection device. }
    \label{fig:4}
\end{figure*}

For a fair experimental training and validation protocol, we built a data-collecting system consisting of a motorized probe system, and captured images during interactions between a probe and an object while recording the interaction forces. As shown in Fig.~\ref{fig:4} (c), we used a RC servo-motor attached to the translation stage for generating the movement. The rod type tool mounted by the translation stage moved up and down (only $z$ direction) automatically to apply force on the object. We measured the interaction force between the tip of the tool and the interaction object through a load cell (model BCL-1L, CAS) and captured the 1280$\times$1024$\times$3 (RGB) images using a 149-Hz camera (Cameleon3, CM3-U3-13Y3C-CS, Pointgrey). We synchronized the collected images and interaction forces using the time stamp of the camera.
%During the interaction between the tip of the tool and the object,
Note that the maximum magnitudes (e.g., 0$N$--12$N$) of the pressing force and pressing time were varied randomly. %concerning a variety of magnitudes of forces and duration of application.

\begin{table}
\caption{The training and test protocols. One set consisted of four touches, with approximately 500 sequential images. The number in the parentheses is the total number of sequential images. The total number of all images was 387,473. }
\label{table:1}
\begin{center}
\begin{tabular}{c|c|c}
\hline
Materials&	Training Set&	Test Set \\
\hline\hline
Sponge	&144 sets (77,097)&	36 sets (19,474)\\
Paper cup	&144 sets (76,966)&	36 sets (19,133)\\
Stapler&	144 sets (77,941)&	36 sets (19,533)\\
Tube&	144 sets (77,849)&	36 sets (19,480)\\
\hline
\end{tabular}
\end{center}
\end{table}

To infer the interaction forces from the images, we selected four objects composed of different materials as shown in Figs.~\ref{fig:4} (a) and (b). Each object had a different rigidity. 
%We collected images of a sponge, a paper cup, a tube, and a stapler. 
We collected images of a sponge, a paper cup, a tube, and stapler. Four target objects selected for this experiment were selected from objects that can be easily obtained around us and that change their appearance largely by the external force. The selected four objects consist of a soft object, a complex object with hard parts, and so on.
To vary the environment around the objects, each object was subjected to four pressing angles ($0^\circ, 10^\circ, 20^\circ, 30^\circ$) and three levels of light intensities (350, 550, 750 lux). Fig.~\ref{fig:4} (a) showed the example images. One image set consists of four contacts to the material, and a total of 15 sets were collected for each environment.
In the end, we collected approximately 380,000 sequential images (e.g., 15 sets$\times$500 images$\times$4 objects$\times$3 lights$\times$4 angles).
We selected three test sets from each material, and the other sets were used to train the deep learning models. Table~\ref{table:1} summarizes detailed information concerning the training and test sets of images of the four objects\footnote{\highlight{The database and the evaluation protocols are released by the following link: https://github.com/hyeon-jo/Interaction-force-estimation-based-on-deep-learning and its code is released by the following link: https://github.com/cxz1418/SSAM\_ForcePrediction}}.

\subsection{Implementation Details}
\begin{table}
\caption{The CNN structure of the baseline method.}
\label{table:2}
\begin{center}
\begin{tabular}{c|c|c}
\hline
Layer Name	&Type&	Size\\
\hline\hline
conv 1/1&	3$\times$3 conv&	16\\
conv 1/2&	3$\times$3 conv&	16\\
\hline
maxpool 1&	stride2	&\\
\hline
conv 2/1&	3$\times$3 conv&	32\\
conv 2/2&	3$\times$3 conv&	32\\
\hline
maxpool 2&	stride2	&\\
\hline
conv 3/1&	3$\times$3 conv&	64\\
conv 3/2&	3$\times$3 conv&	64\\
\hline
maxpool 3&	stride2	&\\
\hline
conv 4/1&	3$\times$3 conv&	128\\
conv 4/2&	3$\times$3 conv&	128\\
\hline
maxpool 4&	stride2	&\\
\hline
conv 5/1&	3$\times$3 conv&	256\\
conv 5/2&	3$\times$3 conv&	256\\
\hline
GAP & & \\
\hline
\end{tabular}
\end{center}
\end{table}

We trained the network weights through the mini-batch stochastic gradient descent by using Adam for 120 epochs. The initial learning rate was le-4, and was multiplied by 1/10 every 30 epochs. In each iteration, a mini-batch of 64 samples was made by sampling 20 sequential training frames, and from each frame, an object was randomly selected. The image then underwent cropping and resizing to a gray-scaled 128$\times$128-pixel image. In the experiment, as a baseline, a variant of the VGG network was used to extract the visual features. As described in Table~\ref{table:2}, the network was composed of 10 layers and output 256 channel feature vectors after the GAP. We also experimented with an 18-layer ResNet \cite{ResNet} to verify that our proposed model works well on other CNNs. To exploit the temporal dynamics, we used the BLSTM network with 256 hidden units and 20 time steps. The last hidden unit feature that was concatenated was fed to 1,024 fully connected layers. Finally, to predict the 1D interaction force, we used the linear-regression model. We trained all models from scratch, and measured performance by using the root mean-squared error (RMSE) and mean absolute error (MAE). We used the MAE as the standard measurement for performance comparisons.

\section{Experimental Results and Discussion}
%\subsection{Experimental Results of Proposed Sequential Attention Module}
\begin{table}
\caption{Experimental results of a baseline, a traditional attention-based method and the proposed multiple frame-based attention method.}
\label{table:3}
\begin{center}
\begin{tabular}{c|c|c|c}
\hline
	Model	&RMSE	&MAE	&Ratio\\
\hline
\hline
	Baseline (CNN+LSTM)&	0.10313&	0.04051&	100\%\\
	\hline
Traditional Spatial&	0.10057&	0.03700&	109\%\\
Traditional Channel	&0.10007&	0.03662&	111\%\\
	Ensemble	&0.09738&	0.03400&	119\%\\
	\hline
Proposed Spatial&	0.09734&	0.03416&	119\%\\
Proposed Channel&	0.09572&	0.03320&	122\%\\
	Ensemble	&0.09356&	0.03183&	127\%\\
\hline
\end{tabular}
\end{center}
\end{table}

In this section, we first provide the overall performance comparison between the baseline~\cite{PAMI2017} and the proposed method. Table~\ref{table:3} shows that the attention methods helped to improve the performance of the baseline by more than 9\% in terms of predicting unknown interaction forces. The channel attention method (or SCAM) was always better than the spatial attention method (or SSAM) in this paper because at high layers of the CNN, high-level features were found in the channel maps of the CNN, and not the spatial maps. Moreover, the proposed ensemble method, by merging the results of the spatial and channel attention methods, effected an improvement of over 27\% over each attention method. Compared with the single frame-based method, with an MAE of 0.0340, the proposed method based on multiple frames always yielded better results, with an MAE of 0.0318 in the ensemble. This indicates that the attention map to infer forces can be effectively generated by exploiting the temporal dynamics of the target object.
%Quantitative evaluation was conducted to find the optimal multi-frame bounds. Table~\ref{Table_frame} shows that the performance improvement was saturated after $n=1$ and in this paper, we use $n=1$ for consideration of the computational complexity and its improvement.
The question that may arise in the next is how many multiple frames are needed as the input images for calculating the saliency attention outputs in the spatial and channel attention methods. In this respect, we conducted quantitative evaluation to find the optimal multi-frames. Table~\ref{Table_frame} shows that the performance improvement was saturated after $n=1$. We eventually use $n=1$ for consideration of the computational complexity and its improvement in this paper.

%\begin{figure}
    %\begin{center}
    %\includegraphics[width=8.0cm]{./pic/fig5.jpg}
    %\end{center}
    %\caption{Results of evaluation according to the number of frames. (a) Mean absolute error. (b) Root mean-squared error. }
    %\label{fig:5}
%\end{figure}

\begin{table}
\caption{The performance changes according to the number of input frames in the proposed method.}
%Results of evaluation according to the number of frames. }
\label{Table_frame}
\begin{center}
\begin{tabular}{c|c|c|c}
\hline
    $n$ & 0 & 1 & 2\\
\hline
\hline
    RMSE & 0.10313& 0.09356&0.09368\\
\hline
    MAE & 0.04051&0.03183 &0.03194\\
\hline
\end{tabular}
\end{center}
\end{table}

We empirically verified that our proposed pooling method is effective at squeezing sequential frame information by comparing two methods of averaging the feature maps: GAP and WAP. From Table~\ref{table:4}, we conclude that the proposed WAP is superior at handling concatenated sequential information and predicting the relevant forces.

\begin{table}
\caption{Performance comparison between GAP and WAP.}
\label{table:4}
\begin{center}
\begin{tabular}{c|c|c|c|c}
\hline
Pooling&	Model	&RMSE	&MAE	&Ratio\\
\hline
	&Baseline	&0.10313&	0.04051&	100\%\\
	\hline
Global &Spatial	&0.09731	&0.03562	&114\%\\
Average&Channel	&0.09599	&0.03431	&118\%\\
Pooling&Ensemble	&0.09411	&0.03311	&122\%\\
	\hline
Weighted &	Spatial	&0.09734	&0.03416&	119\%\\
Average&Channel	&0.09572	&0.03320	&122\%\\
Pooling&Ensemble	&0.09356	&0.03183	&127\%\\
\hline
\end{tabular}
\end{center}
\end{table}

\begin{table}
\caption{Performance comparison of various CNN models.}
\label{table:5}
\begin{center}
\begin{tabular}{c|c|c|c}
\hline
CNN Model	&RMSE&	MAE&	Ratio\\
\hline
	Baseline&	0.10313	&0.04051&	100\%\\
	VGG-like (10 layers)&	0.09356&	0.03183&	127\%\\
	ResNet (18 layers)&	0.09549&	0.03122&	130\%\\
\hline
\end{tabular}
\end{center}
\end{table}

The inference time of the proposed method is measured using PyTorch 1.0 with a single TitanV GPU. As compared in Table~\ref{table:time}, the proposed SSAM and SCAM are slightly slower than the baseline method, respectively. However, the average inference time increase (e.g., 0.03 and 0.034 sec) is not large compared to the accuracy improvement (e.g., 119\% and 122\%). Note that we implement the backbone network architecture without the latest efficient deep learning models such as MobileNet~\cite{MobileNet} and ShuffleNet~\cite{ShuffleNet}. We believe that using this latest computation-efficient network architectures will accelerate the inference time of the proposed method and \cite{Sensors2019} is a good example.  

\begin{table}
\caption{Average inference times are measured with a single TitanV GPU. The data loading time is excluded, and the average inference time is calculated by averaging over 128 times of each method.}
\label{table:time}
\begin{center}
\begin{tabular}{c|c}
\hline
Model & Average Inference Time\\
\hline
Baseline & 0.308 sec\\
\hline
Proposed SSAM & 0.342 sec\\
\hline
Proposed SCAM & 0.338 sec\\
\hline
\end{tabular}
\end{center}
\end{table}

\textbf{Experimental Results on Different Network Architectures} To validate the generality of our method, we applied our model to ResNet \cite{ResNet}, a well-known deep learning architecture. Table~\ref{table:5} shows the comparative results between VGG-like and ResNet. The proposed method worked successfully regardless of the architecture used. For example, the ResNet-based method also achieved results 30\% better in terms of MAE than the baseline.

\begin{table}
\caption{Comparative evaluation with the well-known attention modules.}
\label{table:6}
\begin{center}
\begin{tabular}{c|c|c|c}
\hline
Model	&RMSE	&MAE	&Ratio\\
\hline
Baseline&	0.10313&	0.04051&	100\%\\
SE~\cite{SE}&	0.09838	&0.03769	&107\%\\
CBAM~\cite{CBA}&	0.09974	&0.03745	&108\%\\
Proposed Method&	0.09549&	0.03122&	130\%\\
\hline
\end{tabular}
\end{center}
\end{table}

\textbf{Comparative Evaluation with Well-known Methods} We conducted a comparative analysis with other well-known attention methods. In Table~\ref{table:6}, we provide a summary of the results of the comparative evaluation in terms of inferring the interaction forces on our dataset. The methods tested were our proposed attention module and recently developed state-of-the-art techniques based on the attention mechanism \cite{SE}\cite{CBA}. The proposed method was superior. Note that such methods as in \cite{SE}\cite{CBA} are not designed to generate an attention map from sequential images, and thus suffered performance degradation.

\subsection{Performance Analysis according to Changes in Force Intensity}
\begin{figure}
    \begin{center}
    \includegraphics[width=8cm]{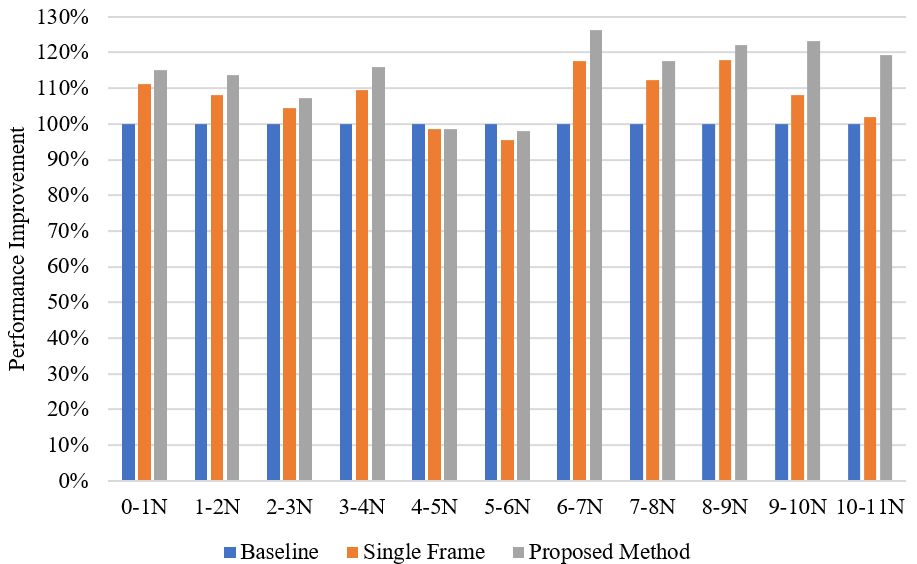}
    \end{center}
    \caption{The results of comparing the MAE of three models: the baseline model, single-frame-based attention model, and our proposed model, using 11 bins according to the magnitude of force.}
    \label{fig:6}
\end{figure}
\begin{figure*}
    \begin{center}
    \includegraphics[width=15cm]{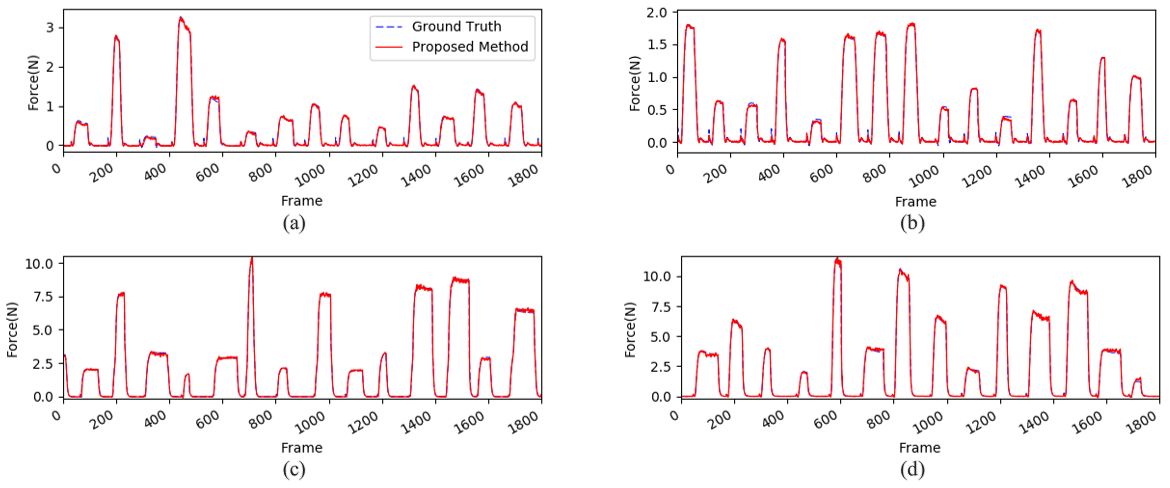}
    \end{center}
    \caption{The results of estimated interaction force using the proposed method on various materials. (a) Sponge, (b) Paper cup, (c) Tube, and (d) Stapler.}
    \label{fig:7}
\end{figure*}
To better understand reasons for why the proposed method improved performance over the baseline method, we divided the magnitudes of the forces into 11 bins, each of which spanned a $1 N$ force interval as shown in Fig.~\ref{fig:6}. We used the MAE measurements of each force interval to determine how the single-frame-based attention method and the proposed method improved compared with the baseline method. From Fig.~\ref{fig:6}, we confirm again that the proposed method of generating attention by using sequential images improved performance in most force intervals.
In detail, from 0 $N$ to 4 $N$, the contact between the tool and the object is initially started, and the interaction force could be measured by the load cell. In this rage, the appearance of the target object begins to change. Since these changes could be concentrated by the attention mechanism, the proposed method helps to improve the accuracy compared with the baseline and the single image-based method. On the contrary, the range from 4 $N$ to 6 $N$ is the interval where the applied force gradually increases, and the performance of the proposed method is saturated.
In relatively strong force intervals, e.g., 9-11 $N$, the proposed method achieved an average improvement of 16\% over the single-frame-based attention model. As the appearance changes of the target object increased when the external force was strong, the proposed method effectively made use of differences in the values of pixels between sequential images to generate attention maps.
%From $0 N$ to $4 N$, the differences in the images were also large, which helped form better attention maps because the tip of the tool touched the target objects in such cases. On the contrary, in 4-6 $N$, the external force was constantly applied to the object, and shape changes in the target were not relatively large. As a result, the attention map was not generated precisely, and performance was slightly worse than for the baseline method.
Overall, the proposed method achieved better performance than the single frame-based attention method.

\subsection{Performance Analysis on Different Objects}
\begin{figure*}
    \begin{center}
    \includegraphics[width=17.5cm]{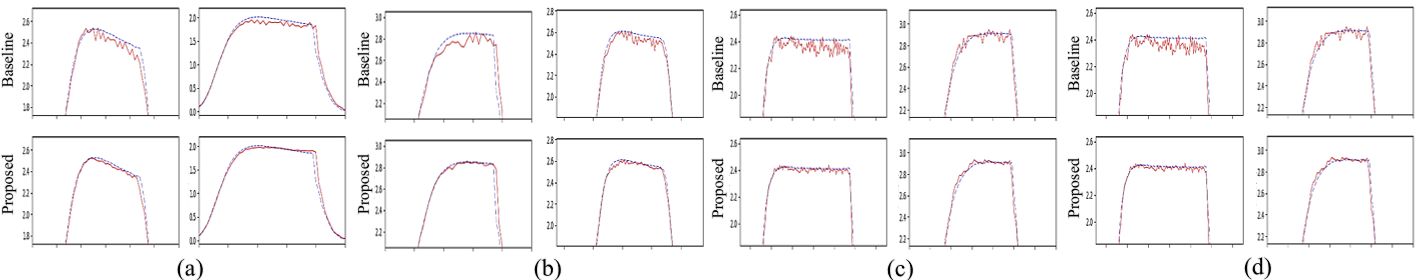}
    \end{center}
    \caption{Comparative prediction results on (a) the sponge, (b) the paper cup, (c) the tube, and (d) the stapler. The blue dotted line means the ground truth and the red thin line is the predicted result. The x-axis and y-axis represent the force ($N$) and the frame, respectively.}
    \label{fig:9}
\end{figure*}

In this section, we investigate the performances according to the different objects. Overall, Fig.~\ref{fig:7} shows that the proposed method predicts the interaction forces using only images, even if the maximum forces are randomly generated.
%Its performance was satisfactory regardless of the object used for experiments.
Looking more closely, Fig.~\ref{fig:9} shows that the proposed method is better than the baseline method when the external force reaches peaks. The baseline method estimated the peak of the interaction force well at first, but its predictions were less stable than those of the proposed method, which were closer to the ground truth. In this respect, we can conclude that temporal dynamics are useful for generating the attention map using the CNN, even though the LSTM make use of the temporal information.

\begin{table}
\caption{The comparative results of improvement rates on the four target objects. }
\label{table:7}
\begin{center}
\begin{tabular}{c|c|c|c|c}
\hline
MAE&	Sponge&	Paper cup&	Tube&	Stapler\\
\hline
\hline
Baseline&	0.02118&	0.02070&	0.06689	&0.05326\\
\hdashline
Ratio (\%)&100\% & 100\% & 100\% &100\%\\
\hline
Single 	&0.01830	&0.01607	&0.06035	&0.04128\\
\hdashline
Ratio (\%)	&116\%	&129\%	&111\%	&129\%\\
\hline
Proposed	&0.01734	&0.01555	&0.05675	&0.03766\\
\hdashline
Ratio (\%)	&122\%	&133\%	&118\%	&141\%\\
\hline
\end{tabular}
\end{center}
\end{table}

\begin{figure}
    \begin{center}
    \includegraphics[width=8.0cm]{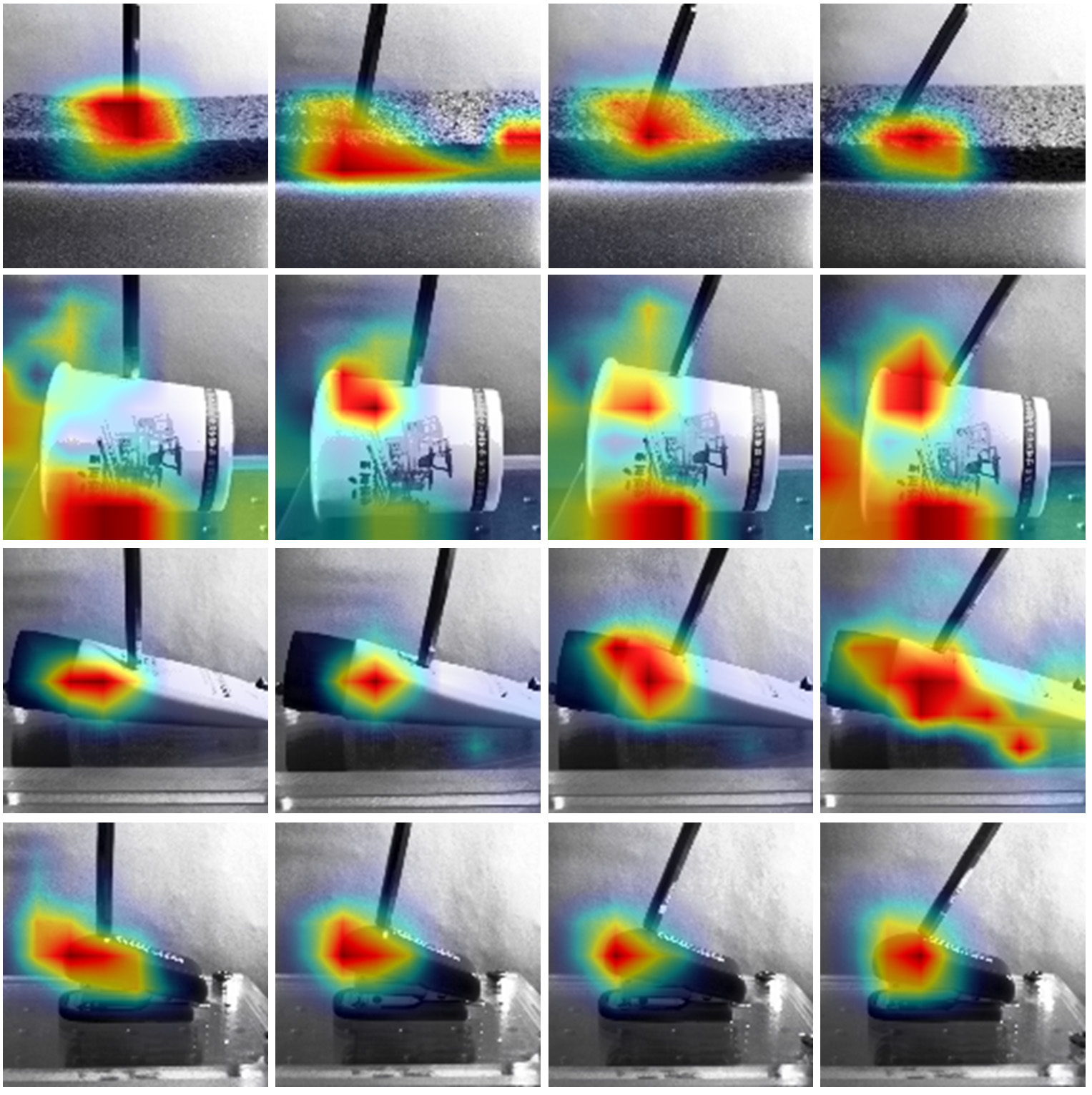}
    \end{center}
    \caption{Example images of the spatial attention map generated by the proposed method. }
    \label{fig:8}
\end{figure}

Table~\ref{table:7} describes the performance improvements according to different target objects and Fig.~\ref{fig:8} illustrates the spatial attention map generated by the proposed method. Specifically, a sponge is an elastic object. Compared with the other objects used, changes to the appearance of the sponge owing to external forces were most apparent, and it thus yielded good results. The proposed method shows the best results on images of the paper cup, as the complex surface textures represent rich visual information. For this reason, it yielded a high estimation accuracy compared with the other rigid objects. As shown in the second row of Fig.~\ref{fig:8}, the network focused on the top and bottom textures of images of the paper cup, where significant changes occurred owing to external forces. The tube was composed of plastic rubbers, and was softer than the other objects, because of which changes to its surface were not obvious. For this reason, the proposed method showed a slightly low improvement on images of the tube.
%In case of a stapler, because it is composed of solid materials, its pattern of shape changes was always constant when an external force was applied.
In case of the stapler, because the stapler has two rigid parts connected around a hinge, when an external force is applied to the upper rigid part of the stapler, its pattern of the shape changes becomes very similar.
In this respect, temporal dynamics played a pivotal role in predicting the interaction forces, this was confirmed by the experimental results in Table~\ref{table:7}. The improvements in the single image-based attention method and the proposed method were 129\% and 141\%, respectively. Compared with the other objects, this 12\% improvement is significant.

\section{Conclusion}
To predict the interaction forces between objects using only images, we developed a sequential image-based attention module that learns a salient model from temporal dynamics. We also proposed a weighted average pooling layer for modifying both spatial and channel attention modules, with the result generated by their ensemble. To verify our method, we collected 359,413 images and information concerning the corresponding interaction forces using an electronic motor-based device. Extensive experiments proved the effectiveness of our method, which achieved better performance than well-known single-frame-based methods. Our proposed method enables the network to concentrate on regions of interaction to infer interaction forces. It serves as good initial research in force prediction using only one vision sensor. In near future, we will release the extended evaluation protocol and corresponding database where we will increase the number of the target objects and the background of the image will be cluttered.

\bibliographystyle{IEEEtrans}
\bibliography{reference}

\begin{IEEEbiography}[{\includegraphics[width=1in,height=1.25in,clip,keepaspectratio]{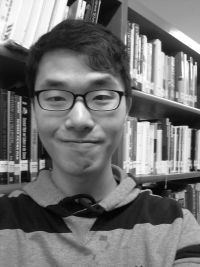}}]{Hochul Shin} received B.S. and M.S. degrees from the Department of Software and Computer Engineering, Ajou University, Korea, in 2017 and 2019, respectively. He is currently a machine learning engineer in NHN Corporation. His research interests include computer vision and gameAI.
\end{IEEEbiography}

\begin{IEEEbiography}[{\includegraphics[width=1in,height=1.25in,clip,keepaspectratio]{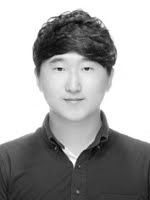}}]{Hyeon Cho} received B.S. degree from the Department of Software and Computer Engineering, Ajou University, Korea, in 2018, and now pursuit Ph.D degree. His recent work is a framework for barcode detection using hand-craft features and an IR-camera based image of a drone. He is currently studying on improving the performance of the action recognition model. His current research interests include computer vision, pattern recognition and deep learning.
\end{IEEEbiography}

\begin{IEEEbiography}[{\includegraphics[width=1in,height=1.25in,clip,keepaspectratio]{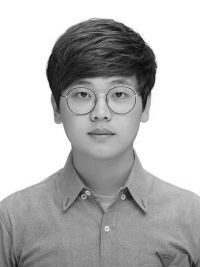}}]{Dongyi Kim} received B.S. and M.S. degrees from the Department of Software and Computer Engineering, Ajou University, Korea, in 2017 and 2019, respectively. His current research interests include computer vision, pattern recognition and deep learning.
\end{IEEEbiography}

\begin{IEEEbiography}[{\includegraphics[width=1in,height=1.25in,clip,keepaspectratio]{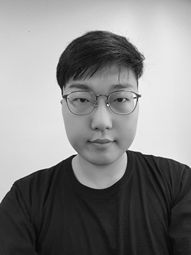}}]{Dae-kwan Ko} received B.S. degree in  Department of Mechanical, Robotics and Energy Engineering, and double degree for Robot Software Convergence from the Dongguk University-Seoul, Republic of Korea, in 2019. His current research interests include robot, haptics, deep learning, and computer vision.
\end{IEEEbiography}

\begin{IEEEbiography}[{\includegraphics[width=1in,height=1.25in,clip,keepaspectratio]{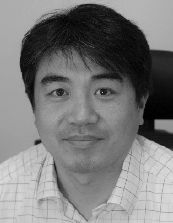}}]{Soo-chul Lim} (M'19) received the B.S., M.S., and Ph.D. degrees in mechanical engineering from the Korea Advanced Institute of Science and Technology, Daejeon, South Korea, in 2001, 2003, and 2011, respectively. From 2006 to 2009, he was a fulltime Lecturer with the Department of Mechanical Engineering, Korea Military Academy. From 2011 to 2016, he was a Research Staff Member with the Samsung Advanced Institute of Technology. In 2016, he joined the Department of Mechanical, Robotics, and Energy Engineering, Dongguk University, Seoul, South Korea, as an Assistant Professor. His current research interests include human–robot interaction, deep learning, surgical robot, and haptics.
\end{IEEEbiography}

\begin{IEEEbiography}[{\includegraphics[width=1in,height=1.25in,clip,keepaspectratio]{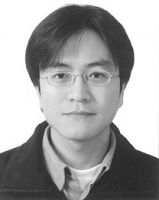}}]{Wonjun Hwang} (M'15) received both B.S. and M.S. degrees from the Department of Electronics Engineering, Korea University, Korea, in 1999 and 2001, respectively, and Ph.D. degree in the School of Electrical Engineering, Korea Advanced Institute of Science and Technology (KAIST), Korea, in 2016. From 2001 to 2008, he was a research staff member in Samsung Advanced Institute of Technology (SAIT), Korea. In 2004, he contributed to the promotion of Advanced Face Descriptor, Samsung and NEC joint proposal, to MPEG-7 international standardization. In 2006, he proposed the SAIT face recognition method which achieved the best accuracy under the uncontrolled illumination situation at Face Recognition Grand Challenge (FRGC) and Face Recognition Vendor Test (FRVT). In 2006, he developed the real-time face recognition engine for the Samsung cellular phone, SGH-V920. From 2009 to 2011, he was a senior engineer in Samsung Electronics, Korea, where he worked on developing face and gesture recognition methods for Samsung humanoid robot, a.k.a RoboRay. In 2011, he rejoined the SAIT as a research staff member and from 2011 to 2014 he worked for a 3D medical image processing of Samsung surgical robot. From 2014 to 2016, he worked on developing deep learning-based face detection and recognition methods for Samsung Galaxy series. In 2016, he joined the department of Software and Computer Engineering, Ajou University, Korea, as an assistant professor. His research interests are in computer vision, pattern recognition, and deep learning.
\end{IEEEbiography}

\EOD

\end{document}